\documentclass[conference]{IEEEtran}

\usepackage[utf8]{inputenc} 
\usepackage[T1]{fontenc}    
\usepackage{hyperref}       
\usepackage{url}            
\usepackage{booktabs}       
\usepackage{nicefrac}       
\usepackage{microtype}      
\usepackage{bookmark}
\usepackage{graphicx}
\usepackage{amsmath,amssymb,amsfonts, bm}
\usepackage{footnote}
\usepackage{subcaption}
\usepackage{siunitx}
\usepackage{mathtools}
\usepackage{hyperref}
\usepackage{tikz}
\usepackage{algorithm}
\usepackage{algorithmic}
\usetikzlibrary{bayesnet}
\usepackage{dblfloatfix}    
\PassOptionsToPackage{bookmarks=false}{hyperref}

\DeclareMathOperator{\Exc}{\mathbb{E}}
\begin{document}

\title{Correlated Mixed Membership Modeling for Somatic Mutations}

\author{\IEEEauthorblockN{Rahul Mehta}
\IEEEauthorblockA{\textit{Department of Bioengineering} \\
\textit{University of Illinois at Chicago}\\
Chicago, IL, USA \\
mehta5@uic.edu}
\and
\IEEEauthorblockN{Muge Karaman}
\IEEEauthorblockA{\textit{Department of Bioengineering}\\
\textit{Center for Magnetic Resonance Research} \\
\textit{University of Illinois at Chicago}\\
Chicago, IL, USA \\
mkaraman@uic.edu}
}

\maketitle

\begin{abstract}
  Recent studies of cancer somatic mutation profiles seek to identify mutations for targeted therapy in personalized medicine.  Analysis of profiles, however, is not trivial, as each profile is heterogeneous and there are multiple confounding factors that influence the cause-and-effect relationships between cancer genes such as cancer (sub)type, biological processes, total number of mutations, and non-linear mutation interactions.  Moreover, cancer is biologically redundant, i.e., distinct mutations can result in the alteration of similar biological processes, so it is important to identify all possible combinatorial sets of mutations for effective patient treatment.  To model this phenomena, we propose the correlated zero-inflated negative binomial process to infer the inherent structure of somatic mutation profiles through latent representations.  This stochastic process takes into account different, yet correlated, co-occurring mutations using profile-specific negative binomial dispersion parameters that are mixed with a correlated beta-Bernoulli process and a probability parameter to model profile heterogeneity. These model parameters are inferred by iterative optimization via amortized and stochastic variational inference using the Pan Cancer dataset from The Cancer Genomic Archive (TCGA).  By examining the the latent space, we identify biologically relevant correlations between somatic mutations.
\end{abstract}


\section{Introduction}\label{intro}

Discerning the relationships between somatic mutations in cancers is the foundation for targeted treatment and patient subtyping.  Since somatic mutations in cancer genomes are often heterogeneous and sparse, where two patients with the same cancer may share only one mutation among thousands, models summarize the high-dimensional interactions into a simpler form.  This requires a model that incorporates multiple confounding variables to determine relationships between somatic mutations.  Based on current literature \cite{van2019identifying, bauer2014cancer} mutually exclusive and co-occurring mutations are influenced by non-linear relationships between gene mutation frequencies, biological processes, cancer (sub)types, total number of mutations in a tumor (TML), and positive selection for mutations.  The combination of multiple confounding variables and the inherent sparsity of somatic mutation data poses a challenge to understand the underlying co-dependencies between mutations.

Statistical and computational models that try to discover relationships between somatic mutations often decompose a patient's mutation profile into a set of higher-level structures that closely resemble known biological processes.  This approach \cite{leiserson2015pan, zhang2016discovery} generally follows a random walk on an existing biological interaction network.  This networks can be modeled as a graph, $G = (V,E)$, where each vertex, $V$ is a gene, and the edge, $E$ denotes the interaction among genes.  The network is then modified into a weighted graph, with edge weights representing probability of interactions and vertex weights corresponding to the frequency of a mutation in a gene.  A walk is then simulated by starting at a mutated gene and moving to another gene based on the probabilities of edge and vertex weights.  The end result is a smaller subnetwork called a functional network that represents an altered biological process.  While functional networks have been validated to discover some aberrant genes and pathways, they often result in false positives due to the inherent assumptions made.

The most common compendium of interaction networks widely used to generate functional networks is the Kyoto Encyclopedia of Genes and Genomes (KEGG) \cite{kanehisa2000kegg}.  The KEGG interaction networks specify genetic pathways, which are complex graphical networks with directed and undirected edges connecting genes based on their physical and biochemical properties.  The genetic pathways are then ascribed to specific biological processes.  For example, the biological process of cell apoptosis (cell death) is controlled by two known genetic pathways compromising of a multitude of different genes.  The networks within the KEGG database, however, are diverse and recapitulate a disease free patient.  Functional networks therefore, assume the interaction networks are also cancer-relevant and disease-specific.  As a result, functional networks are generalized to a common patient population and struggle to discriminate between different cancer types \cite{cowen2017network}.

The second assumption is how functional networks take advantage of mutual exclusivity in somatic mutations.  The process of mutual exclusivity in somatic mutations describes how mutations do not occur together if they are in the same genetic pathway \cite{yeang2008combinatorial}.  In functional networks, accounting for mutual exclusivity corresponds to the frequency of a mutation, which is the weight of a vertex $V_i$ in the graph.  Theory, however suggests that there are multiple confounding factors that cause mutual exclusivity \cite{van2019identifying}.  For example, the mutual exclusivity of mutations in the TP53 and MUC16 genes are better explained by cancer type and TML in cororectal cancer \cite{van2019identifying}.  \par
The last assumption is of preprocessing somatic mutation data to a limited number of mutations.  Although preprocessing is done in numerous studies beyond genetics, there is no gold standard for somatic mutations.  Furthermore, due to the high dimensionalality of somatic mutation data and the limited number of samples, preprocessing will be biased towards frequently occurring mutations \cite{7336349}.  So while a model can identify novel co-occurring mutations within the functional networks, it may only reflect the model's preference for a specific paradigm.  For example, HotNet2 \cite{leiserson2015pan} removes samples with more than 400 somatic mutations, however, overdispersion of a gene and TML are both significant factors that influence the relationship between mutations. \par
From a machine learning perspective we can describe the problems faced by functional networks as overfitting.  This is elucidated by the false positives produced from functional networks.  Specifically, functional networks memorize the parameters of the interaction network instead of learning the parameters of the somatic mutation dataset. So while, functional networks may reproduce valid biological processes, they do not necessarily capture cancer-relevant relationships between mutations.  For example, functional networks validated targeted treatment drugs erlotinib or gefitinib for mutations in the EGFR gene.  Since the EGFR is omnipresent in genetic pathways and as a mutation, functional networks constrain a patient to be mainly influenced by the EGFR mutation.  From the previous example, these treatments, however, are only temporarily effective and a relapse often occurs due to the presence of a co-occurring mutation in the same genetic pathway that had equal cancerous potential \cite{morgillo2016mechanisms}.  So, although functional networks correctly identified the EGFR mutation, it is the interplay between many mutations that influences cancer biology.\par
In this paper, we propose to exploit the inherent latent structure of a somatic mutation dataset with a generative probabilistic model.  Instead of relying on interaction networks \textit{a priori}, we use a prior based on a correlated random measure (CoRM). The CoRM enables the model to specify a notion of similarity on the possible latent distributions of the somatic mutations.  In our case, we would like the latent distributions to correspond to two unique characteristics of the somatic mutation data: mutual exclusivity and cancer-related biological processes.  Specifically, the CoRM assigns probabilities to particular configurations of latent distributions via a Zero Inflated Negative Binomial Process (ZINB) and enforces mutual exclusivity via a correlation structure through the conjunction of the Beta-Bernoulli Process and neural networks. \par
The main contributions of this work are as follows.  We propose the Correlated Zero Inflated Binomial Process, CoZINB, a novel generative latent variable model with an implicit dependency structure and latent parameters that represent the sparsity of a somatic mutation dataset.  Therefore, CoZINB avoids biases from frequency and interaction networks that confound functional network methods.  The CoZINB is used to factorize a somatic mutation profile to infer positively and negatively correlated latent factors\footnote{There are a number of ways to describe sets based on the current nomenclature (clusters, factors, and topics).  For the remainder of the paper we refer to the latent space of our model as factors.} that consist of co-occuring mutations and represent biological processes.  We create a computationally efficient inference scheme with the confluence of stochastic variational inference and amortized variational inference \cite{zhang2019random,ranganath2018correlated,paisley2011discrete}. Our experiments on the pan cancer dataset from The Cancer Genomic Archive (TCGA) verify the benefits of our model to uncover co-occurring mutations while maintaining rules of mutual exclusivity and cancer specific processes.

\section{Background}

\subsection{Negative Binomial Distribution}

The dispersion parameter $r$ of the Negative Binomial (NB) distribution allows models self-excitation \cite{zhou2013negative}, an amenable property in biology, such that if a distinct cancer mutation occurs in a tumor, it is likely to occur again in the tumor.  Another intuitive way to express the NB distribution is to model it as a Poisson-Gamma mixture distribution.  The NB distribution is then a weighted mixture of Poisson distributions, where the rate parameter is an unknown gamma distribution.  Thus, an NB distribution is analogous to an overdispersed Poisson distribution.  Research in differential expression analysis \cite{anders2010differential, dadaneh2018bnp}, regression, and clustering of single-cell data \cite{lopez2018deep} has shown superior performance of NB distribution.  

\subsection{Mixed Membership Modeling}

\begin{algorithm}[h]
    \caption{Latent Dirichlet Allocation a Mixed Membership Model}\label{alg:mm1}
    \begin{algorithmic}
    \FOR{Each cluster $k: 1, ..., K$} 
    \STATE{$\text{Sample } \phi_{k} \sim Dirichlet_{M}(\eta_{1},\dots,\eta_{m})$}
    \ENDFOR
    \FOR{Each sample $j: 1, ..., J$} 
        \STATE $\text{Sample } \theta_{j} \sim Dirichlet_{K}(\alpha_{1},\dots,\alpha_{K})$
        \FOR{Each data point $x_j,m$ in sample $j$}
            \STATE $\text{Sample Cluster } z_{j,m} \sim Discrete_{K}(\theta_j)$
            \STATE $\text{Sample data point } x_{j,m} \sim Discrete_{M}(\phi_{z_{j,m}})$
        \ENDFOR
    \ENDFOR
\end{algorithmic}
\end{algorithm}

\begin{figure*}[t]
  \centering
  \includegraphics[width=0.75\linewidth]{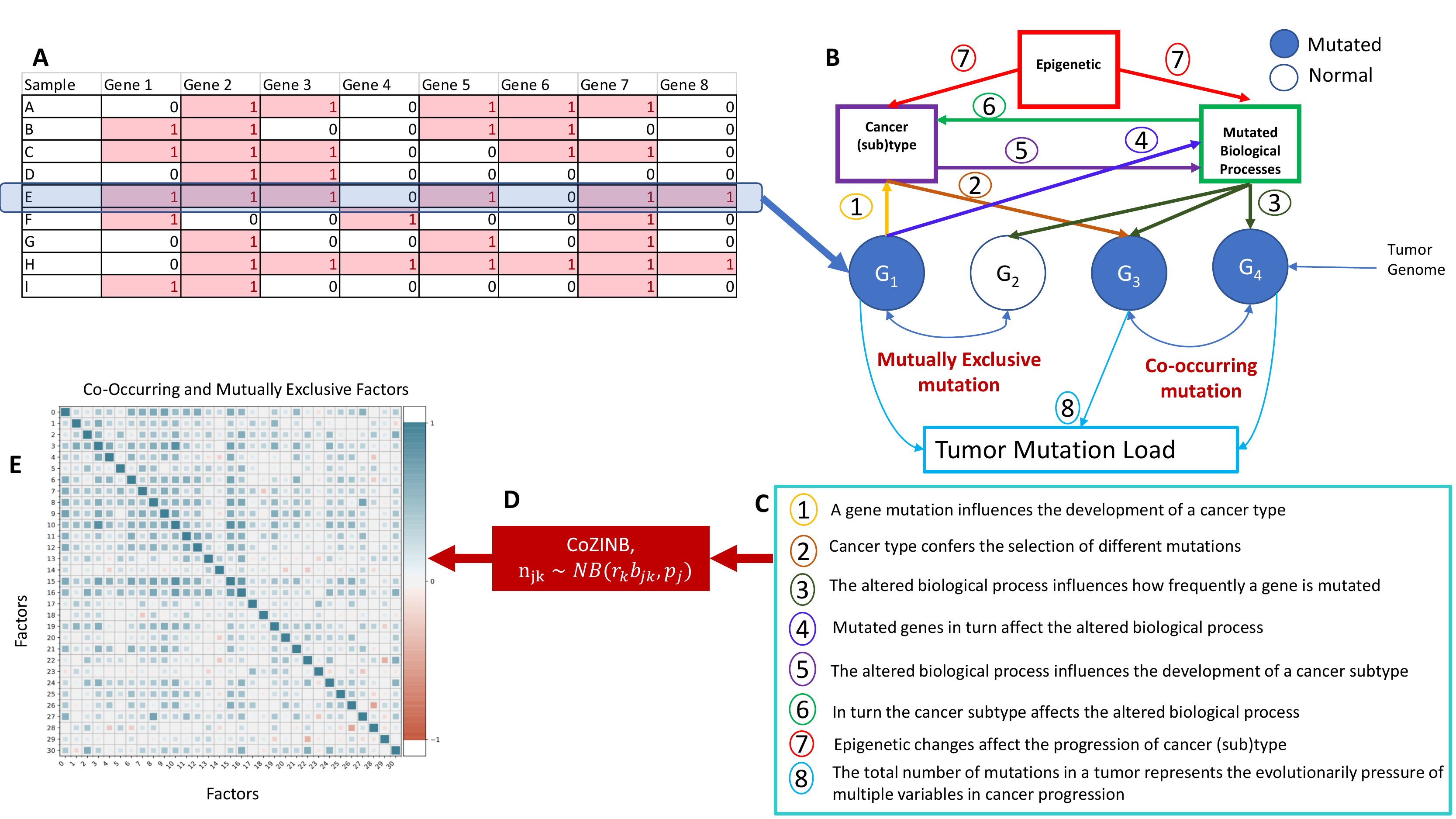}
  \caption{CoZINB models multiple cofounding variables to determine co-occurring mutations and mutually exclusive mutations. (a) Is an example of a set of binarized somatic mutation profiles represented as a matrix, where 1 indicates that a gene is mutated (b-c) There is a complex relationship between cancer, biological processes, and the development of mutations. \cite{van2019identifying} (d) These relationships are modeled via a correlated Beta-Bernoulli Process and an independent Gamma process as seen in Equations \ref{eq:eq2} and \ref{eq:eq4}, where $n_{jk}$ is the number of times the $k_{th}$ factor appears in sample $j$. (e) CoZINB learns a diverse latent structure of somatic mutation profiles, with positively and negatively correlated factors that contain a set of co-occurring mutations}
  \label{fig:flow}
\end{figure*}

Mixed-membership modeling is essentially a soft clustering problem.  In contrast to hard clustering where a data point is assigned to one and only one cluster, each data point in a mixed-membership model is associated with a number of clusters.  It is the interaction among these clusters that gives rise to an observed data point.  In the context of cancer biology, mixed-membership models allow us to remove the restrictive assumption that mutations are mutually exclusive across clusters \cite{liu2010identifying}. 

The generative process of LDA follows Algorithm \ref{alg:mm1}.  What specifically matters, is how choosing the priors: the factor score matrix, $\bm{\theta}$ and the factor loading matrix, $\bm{\phi}$ for different generative models.  For example, if we change the priors in Algorithm \ref{alg:mm1} to Gamma distributions and the likelihood to a Poisson distribution we can obtain Poisson Factor Analysis (PFA) \cite{zhou2013negative}.  Another extension are hierarchical mixed-membership models, which increase model capability by stacking stochastic processes such that algorithm infers the number possible clusters within a dataset \cite{teh2010hierarchical}.

Though powerful, mixed-membership models are limited as they do not explicitly model the correlations between the mixing proportions of any two clusters. Using science articles as an example, the LDA topic model cannot capture that the presence of a topic on cells is more positively correlated with a topic on cancer in comparison to a topic on astronomy.  Recent innovations in joint modeling of correlation and mixture proportions has resulted in improved prediction and interpretability of the topics as seen in \cite{lafferty2006correlated} and an hierarchical extensions in \cite{zhang2019random,paisley2011discrete,ranganath2018correlated}.  Continuing in the context of topic modeling, these methods inject a Gaussian covariance into the $\bm{\theta}$ such that the factor score matrix takes into account the correlation between topics.  Our model follows this format, where the bottom stochastic process is based on correlated random measures.    

\subsection{Latent Space as a Correlated Random Measure} \label{coRM}

Homogenous completely random measure (CRM) \cite{kingman1967completely} is built upon a mean measure (levy measure) as $v(da, dw) = H(da)v_w(dw)$ where $H(da)$ is the base measure and $v_w(dw)$ is the rate measure.  As an example, the Gamma Process ($\Gamma$P) has a mean measure of $v(da, dw) = H(da)e^{-cw}/wdw$.  Intuitively, we can think of the base measure as defining the existence of factors and the rate measure assigning a weight to each factor.  This definition allows us to have an infinite number of factors, but a finite mass, and is a foundation of many of the non-parametric Bayesian statistical models.  

A correlated random measure (CoRM) is created by augmenting a CRM into a higher space to include locations such that the mean measure is now $v(da, dw, dl) = H(da)v_w(dw)v_l(dl)$, where $v_l(dl)$ is a vector of locations in $\mathbb{R}^d$.  We can then draw correlated weights $x$ via a transformation function as $x \sim T(\cdot|w, F(l))$, given the uncorrelated weights, $w$, and a random kernel function on the locations $F(l)$.  Based on Proposition 1 from \cite{ranganath2018correlated} the transformed levy measure is now $\Tilde{v} = v(da, dw, dl)p(dx|F(l),w)$   The transformation distribution, like a homogeneous CRM, is restricted to have finite mass to guarantee generation of useful statistical models.  
\section{Model}

\subsection{Correlated Zero Inflated Negative Binomial Process}

The Correlated Zero-Inflated Negative Binomial Process (CoZINB) is a correlated hierarchical Bayesian model for learning the latent representations of somatic mutation profiles.  We take advantage of neural networks for the random function $F(l)$ to model non-linear correlations of somatic mutations.  The transformation distribution based on the Beta-Bernoulli Process (BeBP) is then used to enforce sparsity of the globally shared latent factors.    We describe this is as a draw from the CoZINB, $X_j \sim NB(RB_{j}, p)$ with a mean measure $\Tilde{v}$: $$ H(da)v_l(dl)\alpha w^{-1}(1-w)^{w-1}p(dx|F(l),w)R_{0}(dr)dw$$
so $B_{j}$ is a draw from the transformation distribution $p(dx|F(l),w)$ and $R$ is an independent $\Gamma$P.   By deriving a latent representation with an unsupervised correlation structure, we use the inherent information shared across the somatic mutation profiles to ensure we capture mutual exclusivity between factors.

As shown in Figure \ref{fig:flow}, by using CoZINB we can model 1) the excessive amount of mutations that do not occur in a patient, 2) non-linear interactions of somatic mutations 3) implicit cancer subtype based on the factors 4) total number of mutations in a somatic mutation profile.  We will show in our experiments the latent factors follow a pattern of co-occurrence that is also observed in cancer biology.

\subsection{Generative Structure}

We represent our dataset as a bag of words commonly seen in text corpus, $X \in \mathbb{N}^{J \times M}$, where J and M are the number of patients and distinct number of mutations, respectively.  As in many factor models we also introduce an additional latent variable, $z_{jm}$, indicating the factor assignment for mutation $m$.  The mutation profile of each patient is realized as a mixture of latent factors ($\phi_{z_{jm}}$) shared by all patients.   Specifically, we model a distinct mutation count in a patient as:
\begin{equation}
\begin{gathered}
\label{eq:eq1}
n_{jmk} = \sum_{i=1}^{N_j} \delta(z_{ji} = k, m_{ji} = m) \\
n_{jk} \sim NB(r_{k}b_{jk}, p_{j}).
\end{gathered}
\end{equation}
From a biological perspective $n_{jk}$ is the number of times a $k$ biological process occurs in patient $j$.  The count $n_{jmk}$ corresponds to the contribution of underlying biological process $k$ to the count of mutation $m$ in sample $j$.  The total number of mutations in patient $j$ is then $N_{j}$.  The shape parameter $r_{k}$ captures the popularity of the biological process $k$ across all patients.  The probability parameter $p_{j}$ models the patient specific somatic mutation profile.  Specifically, $p_{j}$ accounts for heterogeneity among the patient population.  To enable sparsity and correlations within the latent factors, we introduce Bernoulli latent variables $b_{jk}$.  When $b_{jk} = 1$, the latent factor $k$ is present in patient $j$, otherwise irrelevant.  The correlations within the binary random variables are produced via a transformation of the BeBP as discussed in \ref{coRM}.  A transformed distribution follows \cite{ranganath2018correlated} as 
\begin{equation}
\begin{gathered}
b_{jk} \sim Bernoulli(\sigma(\sigma^{-1}(\pi_{k}) + F(\dot))) \\
\pi_k \sim Beta(\alpha/K, \alpha(1 - 1/K))
\end{gathered}
\label{eq:eq2}
\end{equation}
where $\sigma$ is the sigmoid function.  As a result, the proposed model assigns positive probability only to a subset of factors, based on the correlation structured created in $F(\dot)$.

Inspired by Variational Autoencoders (VAE) \cite{kingma2013auto} to model nonlinear correlations we use a deep neural network to create a kernel $F(\dot)$ is:
\begin{equation}
\begin{gathered}
\label{eq:eq3}
f(h_{j}, l_{k}) \sim N(u_{f}(h_{j}, l_{k}), \sigma^{2}_{f}(h_{j}, l_{k})) \\
h_{j} \sim N(0, aI), l_{k} \sim N(0, bI).
\end{gathered}
\end{equation}
where $h_{j}$ is a patient specific vector as the output of an inference network of a VAE.  The locations, $l_{k}$, are then concatenated with $h_{j}$ as in input the decoder of a VAE to create the kernel.  Unlike the standard VAE decoder which aims to recreate the input of a encoder, the decoder in our model supplies prior information to $F(\dot)$.  This paradigm of generating the kernel is similar to Empirical Bayes \cite{carlin2010bayes}, where we use the structure of the data as a prior to generate $F(\dot)$.

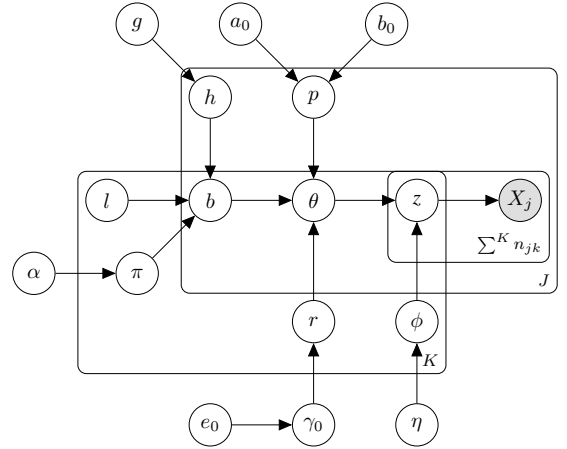
\begin{figure}[h]
\centering
\scalebox{0.8}{
\begin{tikzpicture}
\node[obs] (X) {$X_{j}$};
\node[latent, left=of X] (z) {\textit{z}};
\node[latent, left =of z] (E) {$\theta$};
\node[latent, above =of E] (p) {$p$};
\node[latent, left =of E] (b) {$b$};
\node[latent, below left  =of b] (pi) {$\pi$};
\node[latent, left =of pi] (alpha) {$\alpha$};
\node[latent, above = of b] (h) {$h$};
\node[latent, left = of b] (l) {$l$};
\node[latent, below = 1.3 of E] (r) {$r$};
\node[latent, below =1.3 of z](phik){$\phi$};
\node[latent, below =of phik](eta){$\eta$};
\node[latent, below =of r](smallg){$\gamma_{0}$};
\node[latent, left =of smallg](littleE){$e_{0}$};
\node[latent, above left =of p](ppa){$a_{0}$};
\node[latent, above right =of p](ppb){$b_{0}$};
\node[latent, above left =of h](g){$g$};

\edge {z}{X} ;
\edge{E}{z};
\edge{p}{E};
\edge{h}{b};
\edge{l}{b};
\edge{pi}{b};
\edge{b}{E};
\edge{r}{E};
\edge{phik}{z};
\edge{eta}{phik};
\edge{smallg}{r};
\edge{alpha}{pi};
\edge{g}{h};
\edge{ppa}{p};
\edge{ppb}{p};
\edge{littleE}{smallg}
\plate {pl} {(X)(z)} {$\sum^{K} n_{jk}$};
\plate {pl2} {(pl)(E)(b)(p)} {$J$};
\plate {pl3} {(E)(phik)(b)(pi)(r)(l)} {$K$};

\end{tikzpicture}
}
\caption{Correlated Zero Inflated Negative Binomial Process Topic Model} \label{fig:plate}
\end{figure}

Building upon the above equations, the full generative structure of the CoZINB topic model follows the paradigm of the Gamma-Poisson construction of a NB process:

\begin{equation}
\begin{gathered}
x_{jm} \sim Discrete(\phi_{z_{jm}}),  z_{jm,k} \sim \phi_{mk}\theta_{jk} \\
\phi_k \sim Dirichlet(\eta_{0}\bm{1}_M), N_j = \sum_{k=1}^{K} n_{jk}, \\
 n_{jk} \sim Poisson(\theta_{jk}) \theta_{jk} \sim Gamma(r_{k}b_{jk}, p_{j}), \\
r_k ~ \sim Gamma(\gamma_{0},1/\alpha), \gamma_{0} \sim Gamma(e_{0}, 1/f_{0}), \\
p_{j} \sim Beta(a_{0}, b_{0}), L_{jk} \sim CRT(n_{jk}, r_{k}b_{jk}), \\ L_{k}^{'} \sim CRT(\sum_{j=1}^{J} L_{jk}, \gamma_{0})
\end{gathered}
\label{eq:eq4}
\end{equation}

and $b_{jk}$ follows Equation \ref{eq:eq2}.  To aid in inference of $r_k$, we introduce a data augmentation latent variable, $L_{jk}$ based on the Chinese Restaurant Table as in \cite{zhou2013negative}.  The full model plate is shown in Figure \ref{fig:plate}.  
\subsection{Inference}

Given the set of observed mutation profiles, $X$ and their corresponding mutation counts $m$, our goal is to infer the factors $z_{jm}$, the factor score matrix, $\theta_{jk}$, the factor loading matrix, $\phi_{mk}$, and the factor locations $l_{k}$.  For a biological analogue, $\theta_{jk}$ is the proportion of the biological process $k$ in sample $j$, $\phi_{mk}$ is the proportion of each mutation in factor $k$, and $l_k$ is correlation between a set of co-occurring mutations.   Denote $n_{jk} = \sum^{N_j}\delta(z_{jm})=k$, where $K$ is some finite, but large, truncation level, we posit the fully factorize variational scheme:
\begin{equation}
\begin{gathered}
q(\bm{\theta},\bm{\phi},\bm{\pi},Z,R,P,l,L,h) = \prod^{K} q(l_{k})q(r_{k})q(\pi_{k}) \\
q(\phi_{k}) 
q(\gamma_{0})\prod^{J}\Bigg[ q(h_{j}|X)q(p_{j})\prod^{K}q(b_{jk})q(\theta_{jk})q(L_{jk}) \\ q(L^{'}_{k})\prod^{N_j}_{m=1}q(z_{jm}) \Bigg]
\end{gathered}
\end{equation}
The variational distribution for each latent variable is associated with it's own variational parameter(s) as follows
\begin{equation*}
\begin{gathered}
q(l_{k})= \delta_{\tilde{l_{k}}}, q(\gamma_{0})= \delta_{\tilde{\gamma_{0}}}, \\ 
q(\phi_k) = Dirichlet(\tilde{\eta}) \\
q(\pi_k) = Beta(\tau_{k1},\tau_{k2}) \\ 
q(r_k) = Gamma(\tilde{r_{k1}},\tilde{r_{k2}}) \\
q(p_j) = Beta(\tilde{a_j},\tilde{b_j}) \\
q(b_{jk}) = Bernoulli(\nu_{jk}) \\
q(\theta_{jk}) = Gamma(\tilde{\theta_{jk1}},\tilde{\theta_{jk2}}) \\
q(L_{jk}) = \delta_{\tilde{L_{jk}}} \\ 
q(L^{'}_{k}) = \delta_{\tilde{L^{'}_{k}}} \\ 
q(z_{jm}) = Multinomial(\psi_{jm})
\end{gathered}
\end{equation*}
We let $q(h_{j}|X_{j}) = \delta_{g(X_{j})}$ where $g$ is the inference network of a VAE.  

To update the variational parameters we can use stochastic variational inference (SVI) that assumes we subsample $J_{t} \in J$ patients at every iteration $t$ and optimize the noisy variational objective:
\begin{equation}
\begin{gathered}
L_{t} = \Exc[\ln{p(l,r,\phi,\gamma_{0},\pi)}] + \\ \frac{J}{J_{t}}\sum_{j \in J_{t}}\Exc[ \ln{h_{j},\theta_{j},z_{j},p_{j},b_{j},L_{j},X_{j}}] + H[q(\phi] + \\ \frac{J}{J_{t}}\sum_{j \in J_{t}} H[q(\theta_j,z_j,b_j,p_j)]
\end{gathered}
\end{equation}
At each iteration $t$, we update the local variables, $z, \theta, b,$ and, $p$ using closed form equations, while the remaining variables, with the exception of $\phi$ are updated via stochastic gradient descent via a decreasing step size \cite{hoffman2013stochastic}.  The complete conditional updates are summarized in Table 1.  \footnote{Note that $\Exc[\tilde{\pi_{k2}}] = E[\pi_{k1}](1 - p_{j})^{\Exc[r_{k}]}$.} 

\begin{table*}[h]
\centering
\begin{tabular}{llll}
\hline
\multicolumn{1}{|c|}{\textbf{Latent Variable}} & \multicolumn{1}{c|}{\textbf{Update Type}} & \multicolumn{1}{l|}{\textbf{Variational Update}}                                                                        & \multicolumn{1}{c|}{\textbf{Variational Parameter}} \\ \hline
$\phi_{k}$                                     & Closed                                    & $\eta + \sum_{J \in J_{t}}\sum^{N_j}_{n=1}\psi_{jm}(k)*\bm{1}(X_{jm} = \tilde{m})$                                      & $\eta_{k\tilde{m}}$                                 \\
$\pi_{k}$                                      & Closed                                    & $\alpha/K + \sum^{J}\nu_{jk}$                                                                                           & $\tau_{k1}$                                         \\
$\pi_{k}$                                      & Closed                                    & $\alpha(1-1/K) +J - \sum^{J}\nu_{jk}$                                                                                   & $\tau_{k2}$                                         \\
$r_{k}$                                        & Closed                                    & $\gamma_{0} + \sum^{J}\Exc[L_{jk}]$                                                                             & $r_{k1}$                                            \\
$r_{k}$                                        & Closed                                    & $\alpha - \sum^{J}\Exc[b_{jk}]\ln{1-p_{j}}$                                                                    & $r_{k2}$                                            \\
$p_{j}$                                        & Closed                                    & $a_{0} + N_{j}$                                                                                                         & $\tilde{a}$                                         \\
$p_{j}$                                        & Closed                                    & $b_{0} + \sum^{K} \Exc\left[ r_{k} \right]\Exc\left[ b_{jk} \right]$                                                     & $\tilde{b}$                                         \\
$b_{jk}$                                       & Closed                                    & $\Exc[ \tilde{\pi_{k2}}](\Exc[\tilde{\pi_{k2}}] + 1 - \Exc[ \pi_{k1}])^{-1}$             & $\nu_{jk}$                                          \\
$\theta_{jk}$                                  & Closed                                    & $\Exc\left[ r_{k} \right]\Exc\left[ b_{jk} \right]$ + $\sum^{M}\psi_jm(k)$                                                & $\tilde{\theta_{jk1}}$                              \\
$\theta_{jk}$                                  & Closed                                    & $\Exc\left[\ p_{j} \right]$                                                                                              & $\tilde{\theta_{jk2}}$                              \\
$z_{jm}$                                         & Closed                                    & $\log{\theta_{jk}} + \log{\phi_{km}}$                                                                                   & $\psi_{jm}$                                         \\
$\gamma_{0}$                                         & Gradient Ascent                                    & $\sum^{K}\nabla_{\gamma_{0}}\Exc[p(\gamma_{0})] + \sum^{J}\sum^{K}\nabla_{\gamma_{0}}\Exc\left[ p(L^{'}_{k}|L_{jk}, \gamma_{0}\right)]$                                                                                 & $\delta_{\gamma_{0}}$                                         \\
$l_k$                                          & Gradient Ascent                           & $\sum^{K}\nabla_{l}\Exc\left[\ln{p(l_k)}\right] + \sum^{J}\sum^{K}\nabla_{l}\Exc\left[ p(b_{jk}|\pi_k,h_{n},l_{k}\right]$ & $\delta_{l_k}$                                      \\
$h_n$                                          & Gradient Ascent                           & $\sum^{J}\nabla_{h}\Exc\left[\ln{p(h_j)}\right] + \sum^{J}\sum^{K}\nabla_{h}\Exc\left[ p(b_{jk}|\pi_k,h_{n},l_{k}\right]$ & $\delta_{h_j}$                                      \\
$L_{jk}$                                         & Gradient Ascent                           & $\sum^{J}\sum^{K}\nabla_{L}\Exc\left[ p(L_{jk}|n_{jk},b_{jk},r_{k}\right]$                                               & $\delta_{L_{jk}}$                                  
\end{tabular}
\caption{A summary of the update information for the Variational Parameters.}
\label{tab:my-table}
\end{table*}

\section{Experiments}

To show the importance of modeling sparsity we consider two latent variable models as comparisons:

\textbf{PRME} is a hierarchical mixed membership model based on the Dirichlet Process and the correlations are induced through neural networks with a Gaussian kernel, so $\theta_{jk} \sim Gamma(r_k,f(h_j,l_k))$.  Where $r_k$ is a stick-breaking Dirichlet Process instead of a Gamma Process as in CoZINB.

\textbf{Prod-LDA} \cite{srivastava2017autoencoding} is an extension of LDA to deep generative models.  It replaces the Gaussian Prior in a VAE with a Dirichlet distributed prior to better learn the latent 
structure of text data.

To compare CoZINB's latent factors with functional networks, we use the HotNet2 model.

\textbf{Hotnet2} is used in a discriminate approach to compare functional networks with CoZINB.  We train two distinct linear support vector machine (SVM) classifiers \cite{hearst1998support} with the subnetworks learned by Hotnet2  \cite{leiserson2015pan} and the factors from CoZINB as features.  The SVMs are used to predict the cancer type of a patient given the features.   Additionally we compare the correlation between functional networks and biological process and the correlation between factors from CoZINB and biological processes 

Architecture encoder and decoder follow simple multilayer preceptrons (MLP) with dimensions shown in Table \ref{tab:perp-table}. The hyperparameters for PRME and CoZINB are set as $a = 1, b = 1, \alpha = 1, \eta = 0.2, a_{0} = .001, b_{0} = .001, e_{0} = .001, f_{0} = .001$, $M = 21332$, and a truncation level $K$ as 100 for all models.  All gradient updates are done via Adam \cite{kingma2014adam} with a learning rate of $1e-3$. 

We use the pan cancer dataset from the TCGA consisting of 10296 tumor samples ($J$) and 21332 distinct mutations ($M$).  For training the models we select 70\% of the dataset, setting 20\% for validation of parameters, and 10\% for testing.  The only pre-processing we do is remove 'abParts,' which is not represented in any database.

\subsection{Comparison Metrics}

\textbf{Per-heldout-word perplexity} $$exp(\frac{1}{|X_{Test}|}\sum_{m\in X_{Test}}\ln{p(X_{jm}|X_{j,Train})})$$ is used to measure the utilization of the latent dimensions of the two other latent variable models.

\textbf{Precision@1} measures the influence of the factors on determining the correct count of one distinct (@1) held-out mutation as TML increases.  This is calculated by $$\frac{1}{N}\sum_{1}^{N}\delta(\tilde{X_{jm}}=X_{jm}),$$
where $\tilde{X_{jm}}$ is the predict count of mutation $m$.

\textbf{Specificity} is used to compare the discriminative power of the functional networks and the latent factors of CoZINB in predicting the correct cancer type Lung Adenocarcinoma (LUAD), Colon Cancer (COAD), Ovarianc Cancer (READ), Stomach Cancer, and Breast Cancer (BRCA).

\textbf{Spearman-rho} is used to assess how well the factors produced from CoZINB match with biological processes.  

\textbf{Qualitatively} we analyze the biological implications of the factors obtained by CoZINB for a subset of the TCGA dataset in LUAD and COAD.  We chose these subsets as there are an established number of studies identifying co-occurring and mutual exclusive somatic mutations within these cancers.

\section{Results}
\subsection{Quantitative}
\begin{table}[h]
\begin{tabular}{|c|c|c|c|}
\hline
 &
  Perplexity@H &
  Encoder Sizes &
  Decoder Sizes \\ \hline
Prod-LDA &
  \begin{tabular}[c]{@{}c@{}}@0 1986.30\\@100 1091.33\\ @200 1297.19\\ @400 1691.76\end{tabular} &
  \begin{tabular}[c]{@{}c@{}}$[M \times 100]$\\ $[100 \times 100]$\\ $[100 \times 100]$\end{tabular} &
 \begin{tabular}[c]{@{}c@{}}$[100 \times 100]$\\ $[100 \times 100]$\\ $[100 \times M]$\end{tabular} \\ \hline
PRME &
  \begin{tabular}[c]{@{}c@{}}@0 1151.96\\@100 721.42\\ @200 997.19\\ @400 1091.76\end{tabular} &
  \begin{tabular}[c]{@{}c@{}}$[M \times 1000]$\\ $[1000 \times 1000]$\\ $[1000 x 1000]$\\ $[1000\times40]$\end{tabular} &
  \begin{tabular}[c]{@{}c@{}}$[40\times80]$\\ $[80\times80]$\\ $[80\times80]$\\ $[80\times2]$\end{tabular} \\ \hline
CoZINB &
  \begin{tabular}[c]{@{}c@{}}@0 779.69\\@100 721.77\\ @200 699.35\\ @400 709.34 \end{tabular} &
  \begin{tabular}[c]{@{}c@{}}$[M\times1000]$\\ $[1000\times1000]$\\ $[1000\times1000]$\\ $[1000\times40]$\end{tabular} &
  \begin{tabular}[c]{@{}c@{}}$[40\times80]$\\ $[80\times80]$\\ $[80\times80]$\\ $[80\times2]$\end{tabular} \\ \hline
\end{tabular}
\caption{Perplexity of the held-out test Pan Cancer dataset and architecture of Prod-LDA, PRME, and CoZINB. The @H indicates if the models were trained with only the top frequently occurring mutations i.e., @100 is a training set with only top 100 frequently mutations, @0 = no mutations were held out}
\label{tab:perp-table}
\end{table}
As Table \ref{tab:perp-table} shows, CoZINB performs better than PRME and Prod-LDA when there are a excessive number of zeros in the dataset.  Specifically, we show the importance of using a prior that can incorporate sparsity, by only using the top occurring mutations at levels of 100, 200, and 400 during.  As we limit the amount of mutations the models need to learn, the perplexity of all models begin to improve.   The sparsity of mutations, therefore poses a significant challenge in the learning paradigm.  CoZINB achieved the best performance due to the use of selector variables $b_{jk}$. Specifically the selector variables assigns a finite number of latent factors for each sample and an explicit zero mass to the rest.  Any random signals then will receive a zero probability, wheres the baseline models will assign some small probability to the latent factor resulting in a poor local optima.  

\begin{figure}[h]
  \centering
  \includegraphics[width=.75\linewidth]{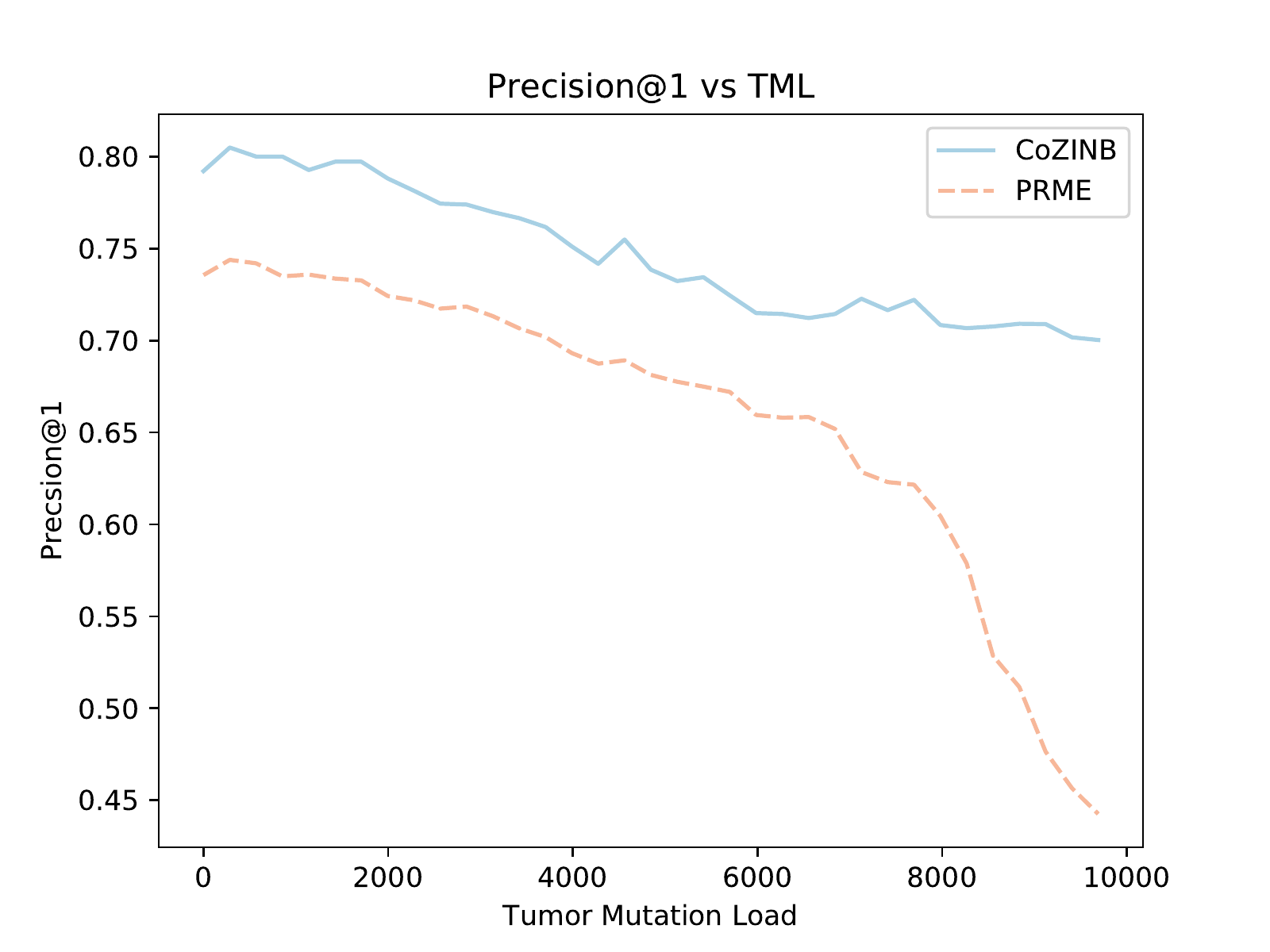}
  \caption{Precision@1 for CoZINB is higher than PRME as it takes account into the sparsity of the dataset through the
  Negative-Binomial process}
  \label{fig:precision}
\end{figure}

In Figure \ref{fig:precision} we observe the influence of the CoZINB shape and probability parameters.  The PRME limits the distribution of the factor score matrix, $\theta_{jk}$, to the frequency of latent factors.  In comparison, CoZINB models the count data as the frequency of the latent factors in $r_k$ and the probability of a count of somatic mutations in $p_j$.  More precisely, the expected count of mutation $m$ in sample $j$ is proportional to $\frac{p_j}{1-p_j}$ and the factor score matrix.

\begin{figure}[h]
  \centering
  \includegraphics[width=.75\linewidth]{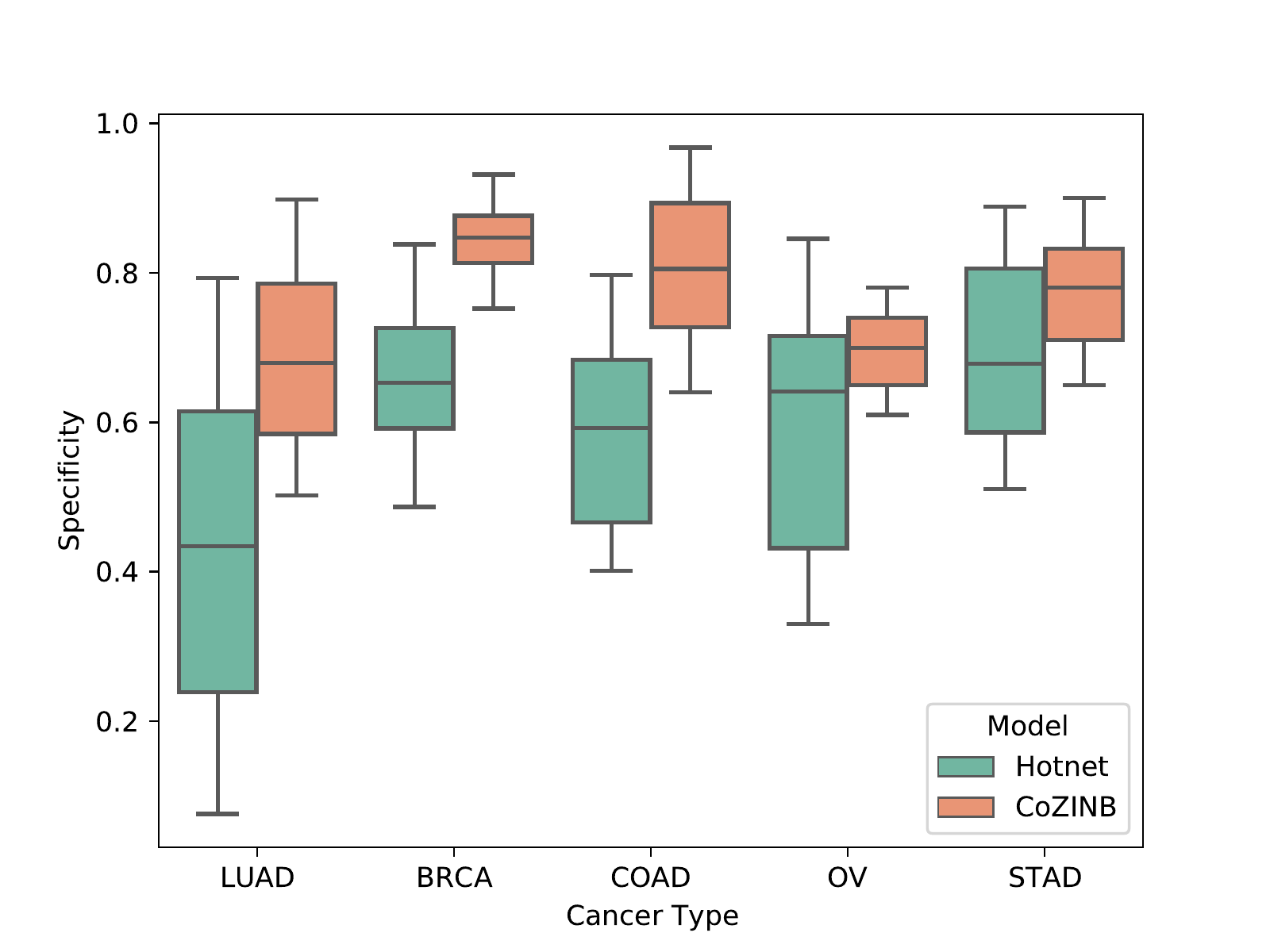}
  \caption{Number of times a factor occurs for all the patients with Lung Adenocarcinoma (LUAD), Colon Cancer (COAD), OV (Ovarian Cancer), STAD (Stomach Cancer),and Breast Cancer (BRCA).}
  \label{fig:hotco}
\end{figure}
\begin{figure}[hb]
  \centering
  \includegraphics[width=\linewidth]{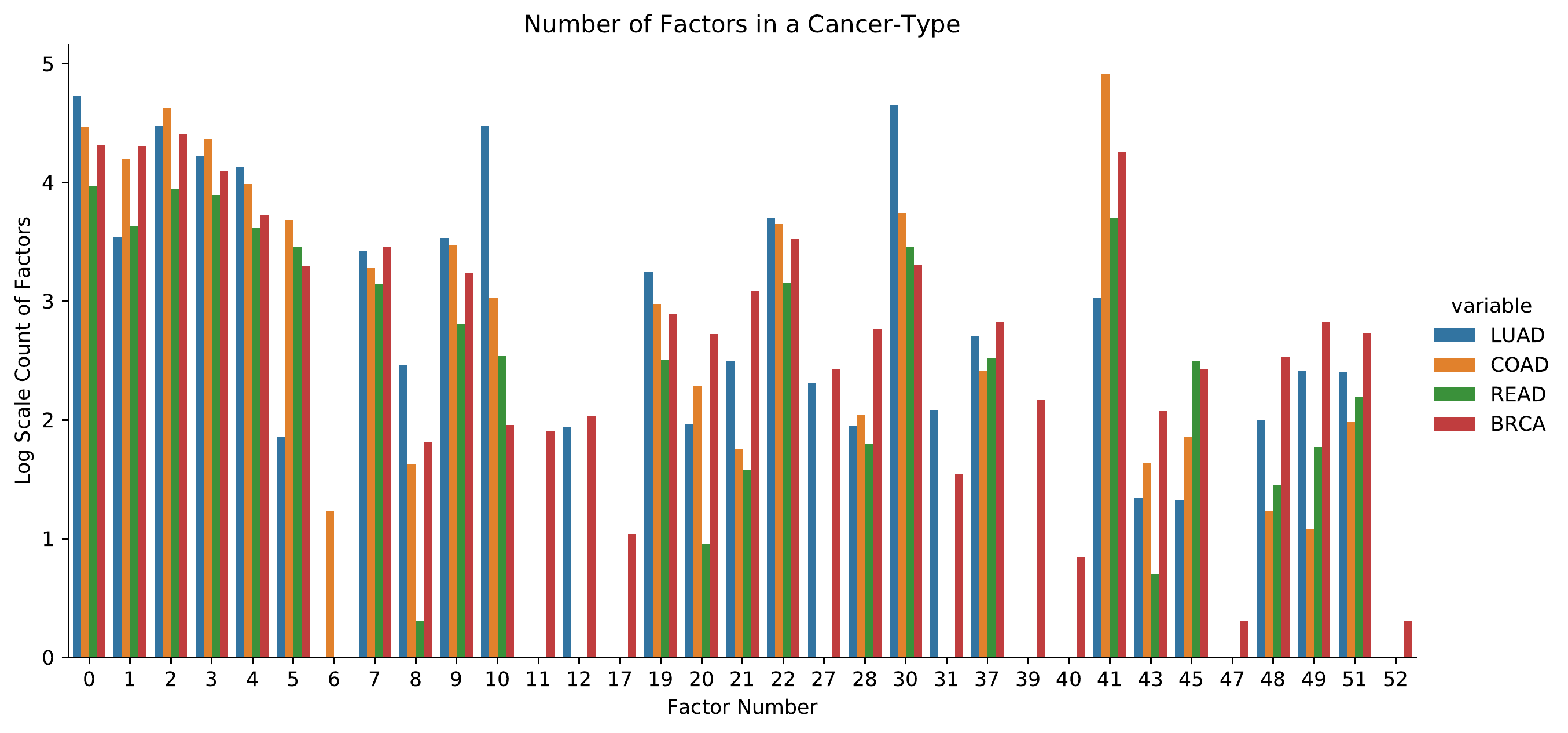}
  \caption{Number of times a factor occurs for all the patients with Lung Adenocarcinoma (LUAD), Colon Cancer (COAD), Renal Cancer (READ), and Breast Cancer (BRCA).}
  \label{fig:combtype}
\end{figure}
Our assertion that functional networks via Hotnet2 are not specific to cancer types and overfit to the structure of interaction networks is shown in Figure \ref{fig:hotco}.  There is a significant variance in predicting cancer type when using the functional networks, especially for LUAD.  The poor performance for HotNet2 in LUAD is likely due to the similarities of mutation profiles of LUAD and LUSC cancers, where LUSC generally has a higher TML.  In contrast, CoZINB factors preserved cancer type information, especially in BRCA where it achieves an average specificity of 0.83.  This suggests that interaction networks, while valuable, have much less influence in determining the relationship between mutations.

In Figure \ref{fig:combtype} we show the interplay of latent factors in LUAD, COAD, READ, and BRCA.  Each bar represents how many times a latent factors occurs in a tissue.  We can see there are distinct factors that only correspond to specific cancers.  We use the spearman-rho coefficient to assess if the latent factors 11, 12, 39, 40, 47, and 52 from BRCA represent a biological process.  We compare it against the spearman-rho coefficient between functional networks of BRCA and biological processes. Using the existing database provided by KEGG, CoZINB has a $\rho$ of 0.67 for the latent factors of BRCA, while HotNet2 is significantly more correlated with a $\rho$of 0.78.  This not unexpected since HotNet2 is built on the prior information from interaction networks.

A key insight from Figures \ref{fig:precision}, \ref{fig:hotco}, and \ref{fig:combtype} is that we observe CoZINB's robust grouping of frequently occurring mutations into a few factors.  The frequently occurring mutations incur a large penalty due to the ZINB process and the optimization procedure prevents the probability mass of frequently occurring mutations to spread across factors.  Thereby, the remaining factors are more diverse and allow for better discriminative power between TML and cancer type.  We see this again in the case studies below, where frequently occurring mutations are grouped into Factors 1-4.

\subsection{Lung Adenocarcinoma Case Study}

A well studied mutual exclusive set of somatic mutations in LUAD are KRAS and EGFR.  They occur in a significant fraction of LUAD patients, but are rarely if-ever observed together in the same tumor.  Since these two genes are in the same pathway and activate similar downstream targets, it is generally assumed that there is no selective pressure to favor cells with both mutations over cells with a mutation in one of them.  This behavior is reproduced in CoZINB, where in the most common factors,  KRAS and EGFR are at the opposite spectrums, with a mean distance of $7185$.  

The factors that are more unique to LUAD, \ref{fig:combtype} in our model are identified by factors 27 with top 5 mutations as: [DNAH5 PCLO ANK3 TTN TP53BP2] and 37 which include the mutations: [ USH2A TTN TSHZ2 DNAH3 MUC16].  To confirm this computationally, we use a simple Random Forest (RF) for classification of cancer type across all patients using the counts the top performing factors as features.  With a significance of $p < 0.05$ RF gives the ranks factors 27 and 37 as the most important features for LUAD, thereby, verifying our model can also implicitly determine patient cancer type.  

\begin{figure}[ht]
  \centering
  \includegraphics[width=\linewidth]{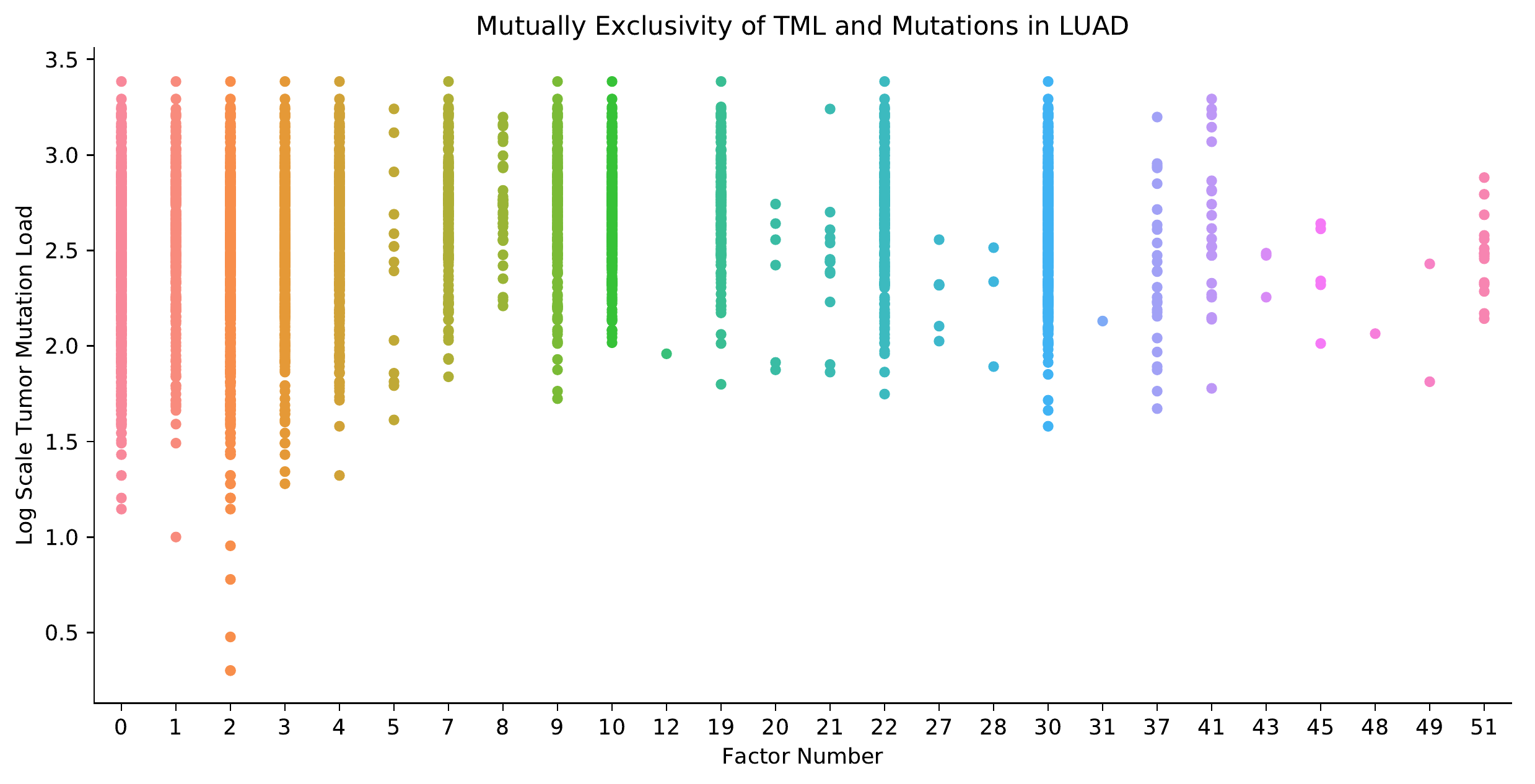}
  \caption{A pattern of mutual exclusivity in LUAD.  The X-axis represents a factor, while the Y-axis the total number of mutations in a tumor.  As tumor mutation load increases, more unique factors appear, specifically ones that are mutually exclusive with certain somatic mutations.}
  \label{fig:luadTML}
\end{figure}

Figure \ref{fig:luadTML} shows the unique factors in every patient with LUAD as TML increases.  Specifically LUAD with high TML will have some form of the background somatic mutations captured in factors 0 to 6, however as, TML begins to increase, occurrences of factors that contain mutually exclusive mutations will appear.   Examining the top underlying somatic mutations of Factor 30 (MUC16, FLG, LRP1B, ZFHX4   CSMD3, RYR) and Factor 1 (PIK3CA, PTEN, TP53, MUC16, MAP3K1,), we observe mutual exclusivity of the MUC15 and TP53 genes as discussed in section \ref{intro}.  
\subsection{Colon Cancer Case Study}
Figure \ref{fig:combtype} shows Factor 6 is unique to the 408 COAD tumor samples within the dataset, with top 5 mutations: [PMVK, BAZ2B, STARD3, SESTD1, KLHL4].  Using the same methodology as before, a RF classifier ranks factors 6 and 41 as the most significant ($p < .05$) features for determining if a patient has Colon Cancer.  Moreover, these specific mutations are also known to commonly occur together as passenger mutations \cite{cerami2012cbio}.

COAD also has specific factors associated with increasing TML in Factors 8 and 21 ($p < .01$) with mutations: [MUC16, DNAH5, TTN, PCLO, ANK3] and [BRCA2, CTNNAL1, DDX52, NAH10, AK9].  Of note are the occurrences of MUC16, PCLO, and BRCA2, which are all hypermutated in colon cancers.  Similar to LUAD we can see Factor 8 is mutually exclusive with factors that include the TP53 mutation with increasing TML.

\begin{figure}[h]
  \centering
  \includegraphics[width=\linewidth]{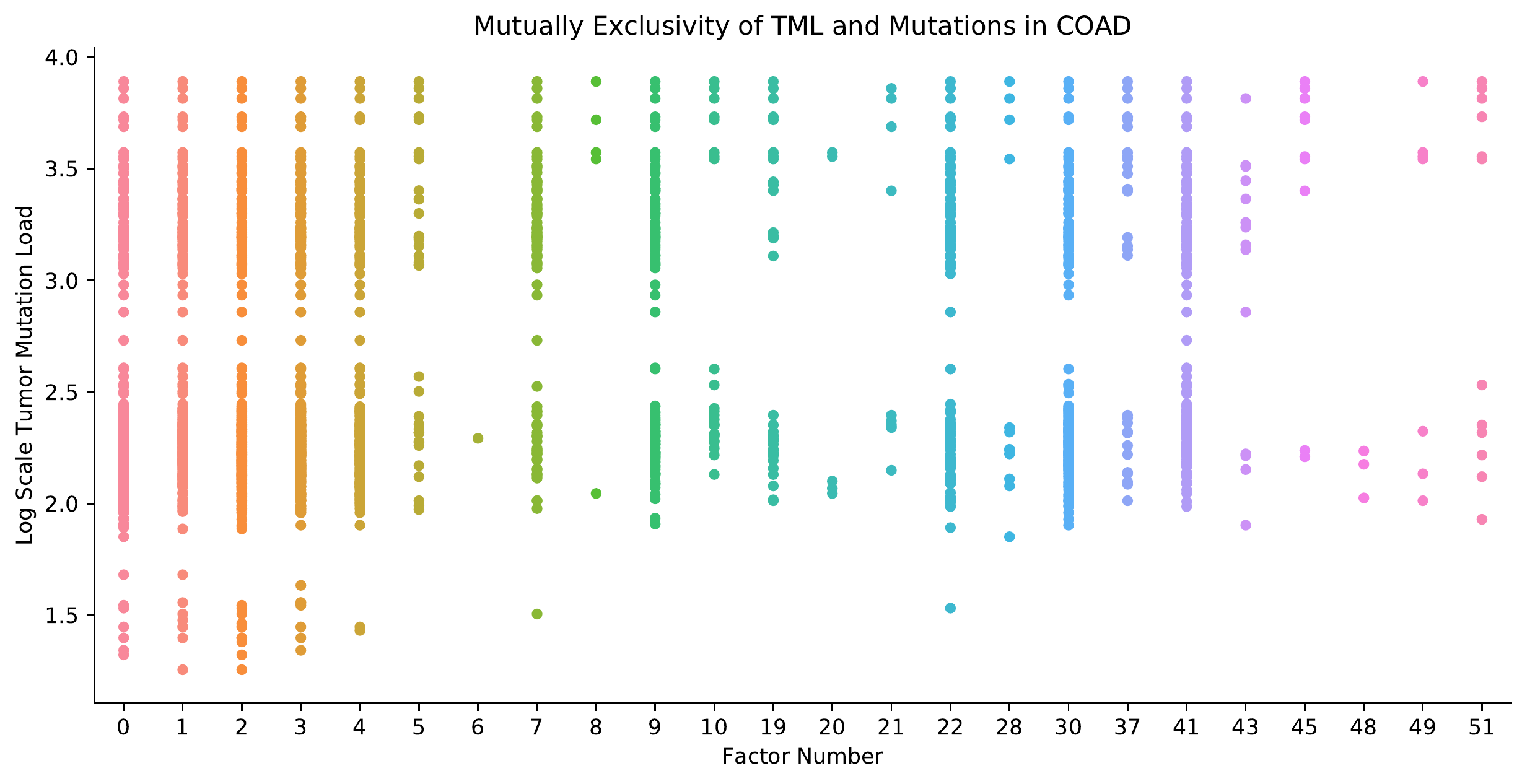}
  \caption{A pattern of mutual exclusivity in COAD.  The X-axis represents a factor, while the Y-axis the total number of mutations in a tumor.  As tumor mutation load increases, more unique factors appear, specifically ones that are mutually exclusive with certain somatic mutations.}
  \label{fig:coadTML}
\end{figure}

\section{Discussion and Future Work}

The CoZINB, unlike existing methods for assessing co-occurring somatic mutations, is unsupervised and infers a latent structure in a sparse and complex dataset.  A major concept shown in our model is that we can probabilistically capture mutual exclusivity  between somatic mutations as a non-linear transformation of multiple latent variables.  We also show that to correctly represent cancer biology and patterns of mutual exclusivity and co-occurrance in somatic mutations, a model needs to incorporate tumor mutational load, cancer type, and non-linear mutation correlations as confounding variables.  

Although, we argued against the use of interaction networks, they might have a place in a semi-supervised framework.  For example, it would be useful to understand the causal relationship of somatic mutations and unaltered genes, analogous to link-prediction in community network detection.  This learning paradigm, however, is a complex combinatoral problem, considering the possible exponential search space of somatic mutation interactions.  

\section*{Acknowledgments}

We thank Aanon Zhang for discussing his work, Population Random Measure Embedding.  We are also grateful for the support of NVIDIA Corporation with the donation of the Titan X Pascal GPU used for this research.  
 
\bibliographystyle{IEEEtran}
\bibliography{mehta5_BIB.bib}

\end{document}